\documentclass[11pt]{article}

\usepackage{acl}

\usepackage{times}
\usepackage{latexsym}
\usepackage[T1]{fontenc}
\usepackage[utf8]{inputenc}
\usepackage{microtype}
\usepackage{inconsolata}
\usepackage{graphicx}
\usepackage{url}
\usepackage{booktabs}
\usepackage{arydshln}
\usepackage{tikz}
\usepackage{pgfplots}
\pgfplotsset{compat=1.18}

\title{ClinicalAligner26AM: A Cross-Lingual Aligner for Dataset Translation; \\ 
Evidences from the MultiClinCorpus Shared Task}

\author{François Remy \\
Parallia Healthcare AI \\
\tt{francois.remy@parallia.eu}}

\begin{document}
\maketitle

\begin{abstract}
Word-level cross-lingual alignment is central to annotation projection, translation auditing, and cross-lingual faithfulness estimation, yet existing neural aligners are rarely adapted to specialized domains.
In this paper, we introduce ClinicalAligner26AM, a large-context multilingual aligner model for biomedical and clinical text initialized from ClinicalEncoder26AM \citep{remy2026clinicalencoder26am}.
Our training recipe is inspired by AWESoME Align \cite{dou-neubig-2021-word}. We build our soft alignment target by sharpening with Sinkhorn--Knop optimal transport a cost matrix established for parallel clinical texts and conversations through the fusion of sentence-level, phrase-level, and token-level signals. We distill this sharpened alignment matrix directly into our student aligner, by encouraging its naive cosine-based token similarity scores to match this target.
At inference time, we project source-span scores through the learned token alignment matrix and decode the longest valid high-scoring span in the target text, optionally supported by MultiClinNER predictions summarized in Appendix~\ref{app:multiclinner-support}.
We evaluate CA26AM on the MultiClinCorpus shared task, which projects Spanish clinical entity annotations into six target languages \citep{multiclinai-overview-2026,smm4h-heard-overview-2026}.
Our two submitted systems ranked respectively first and second across all languages and entity types, with character-weighted F1 scores above 0.95 in nearly all settings.
\end{abstract}

\section{Introduction}
Cross-lingual word alignment has re-emerged as a practical primitive for multilingual NLP \citep{dou-neubig-2021-word,chi-etal-2021-improving}.
When source-language annotations are already available, alignment often offers a more direct and more reliable route to target-language supervision than training a new extractor from scratch \citep{jacqmin-etal-2021-spanalign}.
This is especially true in clinical NLP, where annotated data are scarce, entity boundaries are clinically consequential, and translated documents often preserve content while altering surface form \citep{fraile-navarro-etal-2023-clinical,alhassan-etal-2025-discontinuous}.

The MultiClinCorpus shared task \cite{multiclinai-overview-2026} makes this setting explicit: given Spanish clinical texts with gold entity annotations, the goal is to transfer those annotations into post-edited English, Italian, Dutch, Swedish, Romanian, and Czech translations of these documents.
The central challenge is precise localization: a target mention may expand, contract, reorder, split, or partially lexicalize relative to its source counterpart.
As a result, direct span classification can be unnecessarily rigid, whereas token-level alignment naturally supports boundary variation, discontinuous realizations, nested mentions, and the transfer of span-level metadata.

In this paper, we present ClinicalAligner26AM, a cross-lingual token aligner for biomedical and clinical text derived from ClinicalEncoder26AM \citep{remy2026clinicalencoder26am}.
Our approach is inspired by AWESoME Align \citep{dou-neubig-2021-word}, but adapted to the clinical domain in two key ways.
First, we initialize our student and its teachers with clinically finetuned models, and focus our alignment training on parallel corpora of clinical notes and conversations.
Second, instead of relying on a single bilingual embedding space, we construct a frozen alignment target from multiple complementary signals spanning sentence-level, phrase-level, and token-level semantics, as well as a positional bias.

We also introduce ClinicalAligner26AM-MCAI, a competition variant that injects supervision from manually mapped spans in the MultiClinCorpus training split \cite{lima_lopez_2026_18508039}.

Our contribution is therefore twofold: a domain-specialized multilingual aligner for clinical text, and a simple but effective recipe for training it from frozen multi-view alignment targets.

\begin{figure*}[t]
\centering
\includegraphics[width=\textwidth]{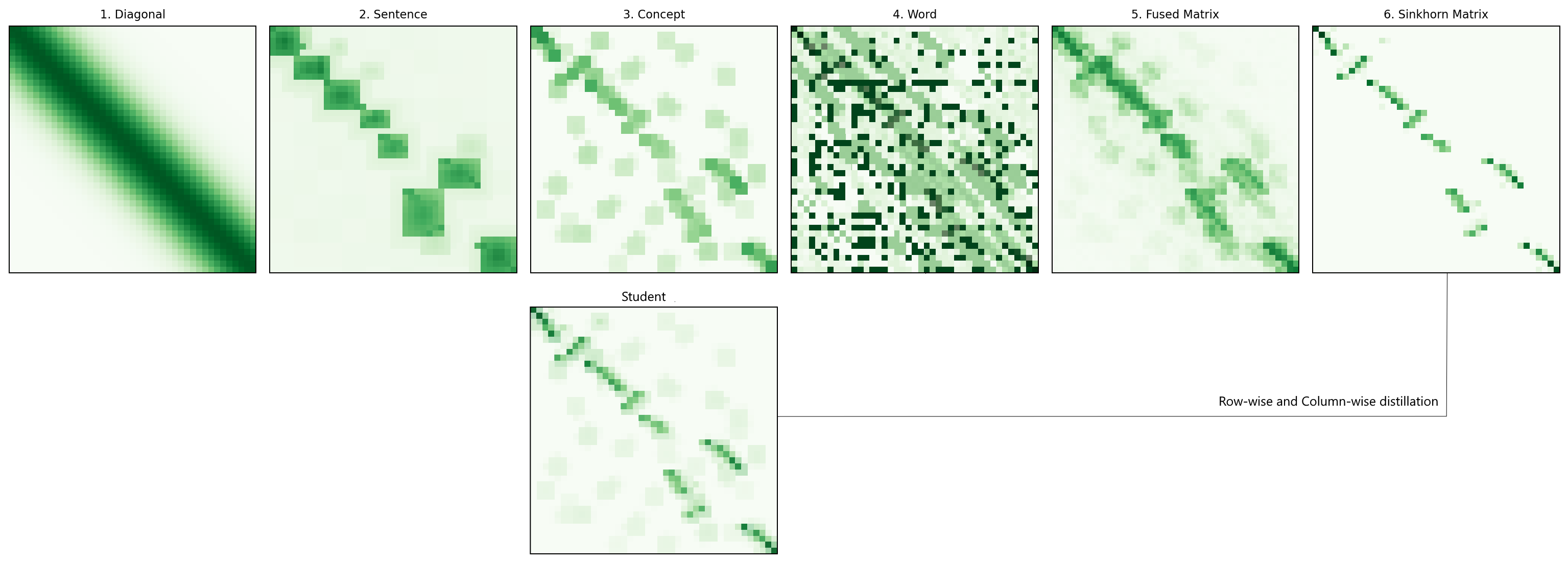}
\caption{Training overview for ClinicalAligner26AM. We combine a diagonal positional prior with sentence-level, concept-level, and word-level similarity matrices to form a fused source--target cost matrix. Sinkhorn--Knopp optimal transport sharpens this fused matrix into a sparse soft alignment target, which is then distilled into the student aligner with row-wise and column-wise supervision.}
\label{fig:training-overview}
\end{figure*}

\section{Methodology}
\vspace{0.25cm}

ClinicalAligner26AM is initialized from ClinicalEncoder26AM and trained as a token-level aligner on parallel clinical texts.
The model is designed so that inference remains simple: source and target texts are encoded independently, their token embeddings are compared with cosine similarity, and the resulting alignment matrix is used directly for annotation projection.
The difficulty lies in training this simple aligner to recover clinically meaningful correspondences without access to paired cross-attention at test time.

Our training recipe therefore separates \emph{teacher construction} from \emph{student learning}, as summarized in Figure~\ref{fig:training-overview}.
On the teacher side, we build a frozen source--target cost matrix that combines a weak positional prior with three complementary semantic views.
The sentence-level view comes from ClinicalEncoder26AM-L2, the phrase-level view comes from ClinicalEncoder26AM-L1, and the token-level view comes from XLM-Align \citep{chi-etal-2021-improving}.
All three teachers are frozen throughout alignment training and never receive gradients.

The positional prior reflects the structure of translated documents, which usually preserve broad discourse order even when local wording changes.
We therefore assign maximum prior mass within a 50-token window around the diagonal, decay this bias up to 100 tokens, and disallow matches beyond that range.
This prior is intentionally weak: it stabilizes long-document matching, but keeps the final alignment driven by semantic compatibility.

\begin{figure*}[t]
\centering
\includegraphics[width=\textwidth]{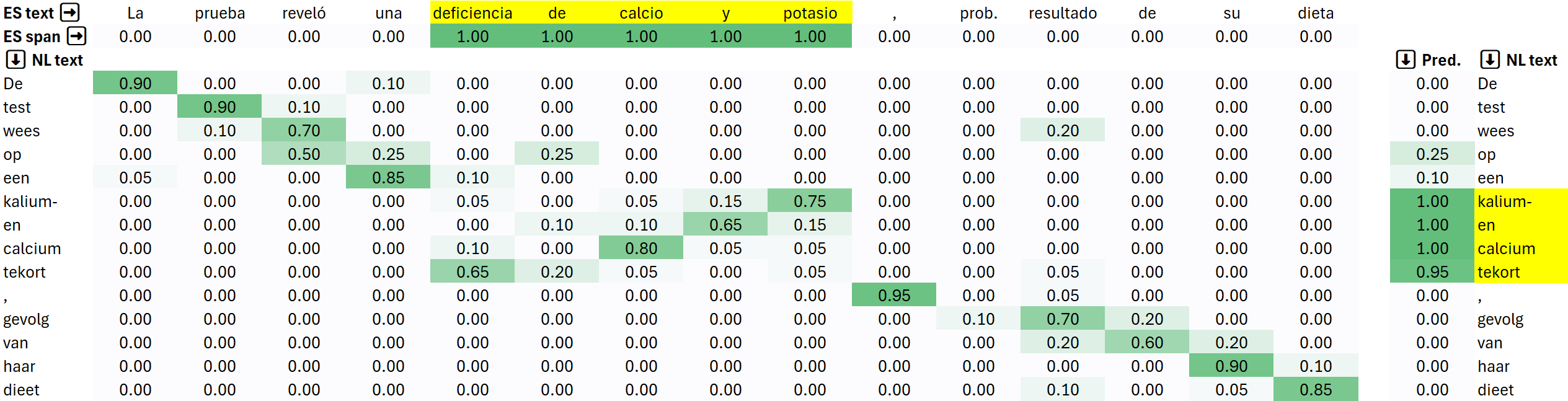}
\caption{Inference overview for span projection. A source-language entity span is represented as token-level membership scores over the source text. The learned alignment matrix projects these scores onto the target tokens, yielding a soft target-side prediction that is then decoded into the final target span, optionally with auxiliary support from the MultiClinNER system summarized in Appendix~\ref{app:multiclinner-support}.}
\label{fig:inference-overview}
\end{figure*}

\newpage
For the token-level view, we found that raw token embeddings are often too noisy for stable clinical alignment.
We therefore compute token similarity with a 5-token contextualization heuristic centered on the token of interest, where the embeddings of the two preceding and the two following token are concatenated to the embedding of the main token, with strong discount factor of 0.1.
This preserves token-centered behavior while reducing spurious matches caused by short tokens carrying little semantic weight.

We then sum the positional and semantic signals into an initial cost matrix and sharpen it with the Sinkhorn--Knopp algorithm (for 100 iterations) to obtain a doubly stochastic soft alignment target.
This optimal-transport step converts heterogeneous similarities into a globally consistent many-to-many alignment distribution.

Our training objective consists of the distillation of this frozen target into ClinicalAligner26AM.
Given independently encoded source and target sequences, the student produces a direct cosine-based alignment matrix.
We train it to match the optimized teacher alignment as closely as possible, in both source-to-target and target-to-source directions. Because the matrix of cross-lingual cosine similarities is inherently not normalized, we rely on a loss combining the row-wise and column-wise cosine similarities between the naive and optimized matrices, instead of KL divergence.
This objective encourages forward and backward consistency while preserving the efficiency of direct embedding comparison at test time.

\newpage
ClinicalAligner26AM-MCAI extends the same framework with additional span-aware supervision from the MultiClinCorpus training split.
During teacher construction, we add a positive bias to source--target token pairs that fall inside manually mapped spans, with the additive scale selected by hyperparameter search.
This keeps the training recipe unchanged at the student level, but nudges the frozen teacher target toward alignments that better reflect the annotation-projection regime of the shared task.

Combining sentence-, phrase-, and token-level signals is particularly useful in translated clinical text.
Sentence-level similarity anchors coarse document structure, phrase-level similarity helps preserve local terminology and paraphrastic variants, and token-level similarity recovers fine-grained boundaries needed for exact span transfer.
The resulting model remains a lightweight token aligner at inference time, but is trained against a substantially richer multi-view notion of cross-lingual correspondence.

\section{Experimental Setup}

We evaluate our systems on the MultiClinCorpus shared task \citep{multiclinai-overview-2026}, where entity annotations from Spanish clinical documents must be projected into their post-edited English, Italian, Dutch, Swedish, Romanian, and Czech translations.
The released data comprise roughly 1{,}200 training documents across the 6+1 languages, together with 250 test documents.
The organizers also provided a background collection of roughly 30{,}000 additional documents for which participants were expected to submit predictions, specifically to make manual correction less practical and to reduce overfitting to the annotated evaluation set.

Following the shared-task protocol, predictions are submitted separately for each entity type (\textsc{Procedure}, \textsc{Disease}, and \textsc{Symptom}) and for each target language.
Our S2 submission corresponds to ClinicalAligner26AM, while S3 corresponds to ClinicalAligner26AM-MCAI.
At inference time, we follow the projection pipeline illustrated in Figure~\ref{fig:inference-overview}.
We first use the learned token alignment matrix to project source-side entity membership scores onto target tokens.
This gives a soft score for each target token indicating how strongly it is supported by the annotated source span.
We then decode a target span from these token scores rather than predicting the span in one step.
Concretely, we choose the longest candidate span whose tokens either exceed a score threshold or are supported by a MultiClinNER prediction, while allowing at most one gap token for punctuation or mis-attributed tokens.
MultiClinNER results are used only as auxiliary support during this decoding stage, not as the primary transfer mechanism; Appendix~\ref{app:multiclinner-support} summarizes that submission and its recall-oriented error profile.

The shared task reports both strict and character-based evaluation metrics, each broken down into precision, recall, and F1.
Strict metrics require an exact match between a predicted span and a gold span, and therefore strongly penalize even small boundary errors.
Character-based metrics instead compare overlap at the character level, giving partial credit when a system identifies the correct mention region but misses the exact start or end boundary.
Because MultiClinCorpus evaluates target-side localization of already annotated source mentions rather than mention discovery from scratch, we treat the character-level metrics as the most informative summary of system quality.

\begin{table*}[t]
\centering
\small
\begin{tabular}{llcccccc}
\toprule
Model Group & Entities & Char R & Char P & Char F1 & Strict R & Strict P & Strict F1 \\
\midrule
\multicolumn{8}{l}{\textit{Aligners}} \\
\textsc{CA26am+mcai} & {Procedure} & $0.96 \pm 0.02$ & $0.96 \pm 0.02$ & $0.96 \pm 0.02$ & $0.83 \pm 0.03$ & $0.82 \pm 0.03$ & $0.82 \pm 0.03$ \\
\textsc{CA26am+mcai} & {Disease} & $0.97 \pm 0.02$ & $0.97 \pm 0.02$ & $0.97 \pm 0.02$ & $0.86 \pm 0.03$ & $0.85 \pm 0.03$ & $0.85 \pm 0.03$ \\
\textsc{CA26am+mcai} & {Symptom} & $0.96 \pm 0.02$ & $0.96 \pm 0.02$ & $0.96 \pm 0.02$ & $0.82 \pm 0.03$ & $0.81 \pm 0.03$ & $0.81 \pm 0.03$ \\
\textsc{CA26am} & {Procedure} & $0.95 \pm 0.02$ & $0.95 \pm 0.02$ & $0.95 \pm 0.02$ & $0.81 \pm 0.03$ & $0.80 \pm 0.03$ & $0.81 \pm 0.03$ \\
\textsc{CA26am} & {Disease} & $0.96 \pm 0.02$ & $0.96 \pm 0.02$ & $0.96 \pm 0.02$ & $0.84 \pm 0.03$ & $0.83 \pm 0.03$ & $0.84 \pm 0.03$ \\
\textsc{CA26am} & {Symptom} & $0.95 \pm 0.02$ & $0.95 \pm 0.02$ & $0.95 \pm 0.02$ & $0.80 \pm 0.03$ & $0.79 \pm 0.03$ & $0.80 \pm 0.03$ \\
\midrule
\multicolumn{8}{l}{\textit{Baseline without source alignment}} \\
\textsc{CE26am} & {Procedure} & $0.85 \pm 0.02$ & $0.81 \pm 0.02$ & $0.83 \pm 0.02$ & $0.74 \pm 0.03$ & $0.70 \pm 0.03$ & $0.72 \pm 0.03$ \\
\textsc{CE26am} & {Disease} & $0.84 \pm 0.03$ & $0.79 \pm 0.02$ & $0.81 \pm 0.02$ & $0.72 \pm 0.03$ & $0.68 \pm 0.03$ & $0.70 \pm 0.03$ \\
\textsc{CE26am} & {Symptom} & $0.79 \pm 0.03$ & $0.73 \pm 0.03$ & $0.76 \pm 0.03$ & $0.65 \pm 0.04$ & $0.60 \pm 0.03$ & $0.62 \pm 0.03$ \\

\bottomrule
\end{tabular}
\caption{Summary of the MultiClinCorpus results of our ClinicalAligner26AM systems (Character-weighted \& Strict Recall, Precision, and F1; all reported as mean $\pm$ standard deviation across the 6 languages).}
\label{tab:main-results}
\end{table*}

\section{Results}

Table~\ref{tab:main-results} summarizes the main outcome of the shared task.
Our two aligner submissions ranked first and second overall across all target languages and all three entity types, with very small gaps between them.
Character-level F1 is above 0.95 in most settings, and approaches 0.97 for several language--entity combinations, especially in English, Italian, and Romanian.
These results show that token-level alignment is sufficient to recover highly accurate target spans once source annotations are provided, when an appropriate long-context in-domain neural aligner is available.

The comparison with the CE26AM baseline is particularly instructive.
In our earlier MultiClinNER setting, performance was limited by the usual difficulty of mention discovery, which created visible precision issues and a larger recall--precision trade-off.
Here, the source entities are already known, so the task becomes one of localization rather than detection.
This substantially simplifies the problem and explains why the aligner models improve so sharply over the NER-only baseline, especially on the character-based metrics that best reflect near-miss boundary errors.
For completeness, Appendix~\ref{app:multiclinner-support} includes a compact summary of that earlier submission, including its ranking table, optimization curve, and error profile.

The MCAI variant generally improves over the base aligner, but only modestly.
This small gap is itself informative: most of the performance appears to come from the quality of the frozen multi-view alignment target rather than from task-specific span supervision alone.
At the same time, the consistent edge of CA26AM+MCAI suggests that annotation-aware biasing still helps refine difficult boundary decisions.

More broadly, these results support token-level alignment as an effective mechanism for transferring exact clinical spans across translated documents.
Because the method operates at the token level, it can in principle support split mentions, nested annotations, and metadata transfer even though MultiClinCorpus does not fully stress these capabilities.
This same property also makes the model attractive for downstream applications such as translation auditing, error localization, and cross-lingual faithfulness assessment.

\section{Conclusion}

In this work, we introduced ClinicalAligner26AM, a ClinicalEncoder26AM-initialized cross-lingual token aligner for biomedical and clinical text.
Our central idea is to train a simple inference-time aligner from a richer frozen teacher target built from multiple semantic views and sharpened with optimal transport.
We further presented ClinicalAligner26AM-MCAI, which injects annotation-aware alignment bias from the MultiClinCorpus training split.

Across the MultiClinCorpus shared task, the two variants ranked first and second overall and achieved near-perfect multilingual entity transfer, with character-level F1 above 0.95 in nearly all settings.
These results suggest that when source-language annotations are already trusted, token-level alignment is an especially effective way to turn cross-lingual transfer into a localization problem rather than a full mention-discovery problem.

More broadly, the same formulation should extend naturally to richer annotation schemas, including nested spans and metadata transfer, and to downstream settings such as translation auditing, document comparison, and summary assessment.

\newpage
\bibliography{custom}

\clearpage
\onecolumn
\appendix

\section{Appendix: MultiClinCorpus Full Results}

\begin{table}[h!]
\centering
\small
\begin{tabular}{llcccccc}
\toprule
& & \multicolumn{2}{c}{CA26AM+MCAI} & \multicolumn{2}{c}{CA26AM} & \multicolumn{2}{c}{CE26AM} \\
Entity Type & Lang & Char F1 & Strict F1 & Char F1 & Strict F1 & Char F1 & Strict F1 \\
\midrule
PROCEDURE & cz & 0.946 & 0.819 & 0.935 & 0.806 & 0.828 & 0.688 \\
PROCEDURE & en & 0.961 & 0.841 & 0.958 & 0.840 & 0.835 & 0.711 \\
PROCEDURE & it & 0.963 & 0.799 & 0.958 & 0.771 & 0.828 & 0.697 \\
PROCEDURE & nl & 0.916 & 0.770 & 0.907 & 0.755 & 0.811 & 0.692 \\
PROCEDURE & ro & 0.967 & 0.856 & 0.965 & 0.849 & 0.840 & 0.722 \\
PROCEDURE & sv & 0.946 & 0.806 & 0.938 & 0.795 & 0.828 & 0.702 \\
\midrule
DISEASE & cz & 0.957 & 0.851 & 0.941 & 0.824 & 0.798 & 0.674 \\
DISEASE & en & 0.976 & 0.896 & 0.968 & 0.884 & 0.851 & 0.752 \\
DISEASE & it & 0.975 & 0.882 & 0.966 & 0.843 & 0.824 & 0.703 \\
DISEASE & nl & 0.932 & 0.820 & 0.924 & 0.804 & 0.793 & 0.678 \\
DISEASE & ro & 0.974 & 0.898 & 0.967 & 0.878 & 0.831 & 0.719 \\
DISEASE & sv & 0.943 & 0.828 & 0.946 & 0.825 & 0.806 & 0.681 \\
\midrule
SYMPTOM & cz & 0.951 & 0.811 & 0.936 & 0.791 & 0.736 & 0.585 \\
SYMPTOM & en & 0.971 & 0.879 & 0.964 & 0.864 & 0.779 & 0.654 \\
SYMPTOM & it & 0.972 & 0.831 & 0.962 & 0.787 & 0.764 & 0.598 \\
SYMPTOM & nl & 0.921 & 0.768 & 0.915 & 0.763 & 0.726 & 0.564 \\
SYMPTOM & ro & 0.971 & 0.861 & 0.962 & 0.834 & 0.774 & 0.625 \\
SYMPTOM & sv & 0.954 & 0.811 & 0.947 & 0.799 & 0.767 & 0.619 \\
\bottomrule
\end{tabular}
\caption{Full leaderboard summary for the Parallia submission across entity types and languages, reporting the Character-weighted and Strict F1 scores for our two aligner models (ClinicalAligner26AM-MCAI and ClinicalAligner26AM) as well as our NER-only baseline (ClinicalEncoder26AM) detailled in Annex \ref{app:multiclinner-support}.}
\label{tab:appendix-full-results}
\end{table}

\section{Appendix: MultiClinNER Support Submission}
\label{app:multiclinner-support}

\newcommand{\rankmedalicon}[3]{
  \tikz[baseline=-0.6ex, x=1ex, y=1ex]{
    \fill[blue!70!black] (-0.60,1.00) -- (-0.18,0.30) -- (-0.42,0.30) -- (-0.82,1.00) -- cycle;
    \fill[blue!50!cyan] (-0.42,1.00) -- (-0.05,0.30) -- (0.08,0.30) -- (-0.02,1.00) -- cycle;
    \fill[blue!70!black] (0.60,1.00) -- (0.18,0.30) -- (0.42,0.30) -- (0.82,1.00) -- cycle;
    \fill[blue!50!cyan] (0.42,1.00) -- (0.05,0.30) -- (-0.08,0.30) -- (0.02,1.00) -- cycle;
    \fill[#1] (0,0) circle (0.75);
    \draw[black!55, line width=0.08ex] (0,0) circle (0.75);
    \node[font=\bfseries#3, text=white] at (0,0) {#2};
  }
}
\newcommand{\rankmedal}[1]{
  \makebox[2.4em][c]{
    \ifnum#1=1
      \rankmedalicon{orange!80!yellow}{1}{\fontsize{5}{5}\selectfont}
    \else\ifnum#1=2
      \rankmedalicon{black!30}{2}{\fontsize{5}{5}\selectfont}
    \else\ifnum#1<6
      \rankmedalicon{brown!70!orange}{#1}{\fontsize{5}{5}\selectfont}
    \else\ifnum#1<10
      \rankmedalicon{black}{#1}{\fontsize{5}{5}\selectfont}
    \else
      \rankmedalicon{black}{#1}{\fontsize{5}{5}\selectfont}
    \fi\fi\fi\fi
  }
}
\newcommand{\rank}[1][]{
  \rankmedal{#1}
}

This appendix summarizes the MultiClinNER submission whose predictions are optionally used during CA26AM decoding. These predictions performed particularly well in the leaderboard, especially in recall.

\begin{table}[h!]
\centering
\scriptsize
\begin{tabular}{llcccccccc}
\toprule
Entity Type & Lang & Char R rk & Char P rk & Char F1 rk & Strict R rk & Strict P rk & Strict F1 rk & Char F1 & Strict F1 \\
\midrule
PROCEDURE & cz & \rank[1] & \rank[8] & \rank[3] & \rank[3] & \rank[7] & \rank[5] & 0.8276 & 0.6880 \\
PROCEDURE & en & \rank[1] & \rank[11] & \rank[5] & \rank[4] & \rank[11] & \rank[8] & 0.8348 & 0.7110 \\
PROCEDURE & es & \rank[2] & \rank[11] & \rank[8] & \rank[8] & \rank[11] & \rank[8] & 0.8476 & 0.7536 \\
PROCEDURE & it & \rank[1] & \rank[9] & \rank[3] & \rank[2] & \rank[8] & \rank[5] & 0.8275 & 0.6969 \\
PROCEDURE & nl & \rank[2] & \rank[10] & \rank[3] & \rank[3] & \rank[10] & \rank[5] & 0.8112 & 0.6922 \\
PROCEDURE & ro & \rank[1] & \rank[9] & \rank[3] & \rank[2] & \rank[8] & \rank[4] & 0.8396 & 0.7222 \\
PROCEDURE & sv & \rank[1] & \rank[8] & \rank[3] & \rank[3] & \rank[8] & \rank[5] & 0.8275 & 0.7015 \\
\midrule
DISEASE & cz & \rank[1] & \rank[10] & \rank[4] & \rank[3] & \rank[8] & \rank[5] & 0.7983 & 0.6737 \\
DISEASE & en & \rank[1] & \rank[13] & \rank[6] & \rank[7] & \rank[13] & \rank[9] & 0.8510 & 0.7520 \\
DISEASE & es & \rank[6] & \rank[10] & \rank[8] & \rank[8] & \rank[9] & \rank[8] & 0.8431 & 0.7512 \\
DISEASE & it & \rank[1] & \rank[11] & \rank[3] & \rank[3] & \rank[9] & \rank[8] & 0.8235 & 0.7027 \\
DISEASE & nl & \rank[3] & \rank[8] & \rank[3] & \rank[3] & \rank[9] & \rank[7] & 0.7926 & 0.6778 \\
DISEASE & ro & \rank[1] & \rank[10] & \rank[3] & \rank[3] & \rank[6] & \rank[3] & 0.8307 & 0.7189 \\
DISEASE & sv & \rank[1] & \rank[8] & \rank[4] & \rank[3] & \rank[6] & \rank[5] & 0.8058 & 0.6809 \\
\midrule
SYMPTOM & cz & \rank[1] & \rank[12] & \rank[6] & \rank[4] & \rank[12] & \rank[9] & 0.7360 & 0.5847 \\
SYMPTOM & en & \rank[1] & \rank[13] & \rank[8] & \rank[8] & \rank[14] & \rank[11] & 0.7791 & 0.6542 \\
SYMPTOM & es & \rank[3] & \rank[9] & \rank[6] & \rank[6] & \rank[9] & \rank[6] & 0.7914 & 0.6696 \\
SYMPTOM & it & \rank[1] & \rank[12] & \rank[5] & \rank[3] & \rank[12] & \rank[9] & 0.7637 & 0.5976 \\
SYMPTOM & nl & \rank[1] & \rank[15] & \rank[3] & \rank[4] & \rank[16] & \rank[9] & 0.7258 & 0.5637 \\
SYMPTOM & ro & \rank[1] & \rank[10] & \rank[3] & \rank[3] & \rank[9] & \rank[6] & 0.7743 & 0.6248 \\
SYMPTOM & sv & \rank[1] & \rank[10] & \rank[3] & \rank[3] & \rank[9] & \rank[5] & 0.7666 & 0.6187 \\
\bottomrule
\end{tabular}
\caption{Full leaderboard summary for the MultiClinNER submission based on ClinicalEncoder26AM. The medal-style rank display is intentionally preserved from the original system paper.}
\label{tab:appendix-multiclinner-results}
\end{table}

\noindent{}In the next page, we summarize the pre-training, NER finetuning of our ClinicalEncoder26AM model. We also provide a detailed analysis of our results in Table \ref{tab:appendix-multiclinner-results}, and refer to \citet{remy2026clinicalencoder26am} for more details.

\twocolumn

\paragraph{Pretraining.}
ClinicalEncoder26AM is a clinically post-trained multilingual encoder built on top of BGE-M3 \citep{chen-etal-2024-m3} and aligned with a clinically relevant latent space at multiple levels, including sentence-, context-, and concept-level semantics, following the Diagnosable ColBERT methodology \citep{diagnosable_colbert_2026}.
Its backbone follows an XLM-RoBERTa-style architecture \citep{conneau-etal-2020-unsupervised} with 24 Transformer layers, a hidden size of 1024, 16 attention heads, and support for up to 8192 tokens in a single pass.
Its post-training combines synthetic clinical notes, patient--doctor conversations, curated translations, and annotated biomedical supervision such as MedMentions \citep{mohan-li-2019-medmentions}.
During this stage, supervision is applied at multiple granularities, including a ColBERT-style late-interaction objective.
This multi-view adaptation is designed to preserve the multilingual alignment inherited from BGE-M3 while making token embeddings more clinically meaningful and diagnosable.
Because the encoder can project tokens into ontology-linked concepts such as UMLS \citep{bodenreider2004umls} and SNOMED CT \citep{donnelly2006snomedct}, it offers broad semantic coverage that is useful both for mention linking and for identifying clinically plausible evidence spans.

\paragraph{Methodology.}
For the MultiClinNER shared task, ClinicalEncoder26AM is finetuned as a lightweight BIO tagger for procedure, disease, and symptom extraction.
Indeed, while good semantic linking and good boundary detection are related, they are not the same problem.
A clinically grounded encoder may already place tokens near the right ontology concepts or clinically similar contexts, yet exact span extraction still requires deciding where a mention begins and ends in free text.
For that reason, the original system uses a simple supervised tagging head on top of the pretrained encoder rather than relying on ontology projection alone.
More specifically, the encoder outputs are passed through a two-layer CNN block with window size 5 and hidden size 1024 before the final token-classification layer.
This CNN design is intended to capture the short-range contextual cues that are especially important for span boundary detection, while keeping the overall tagging architecture lightweight.
In practice, most documents fit in a single long-context forward pass, and this CNN-augmented tagging head learns local BIO transitions on top of semantically rich token representations.

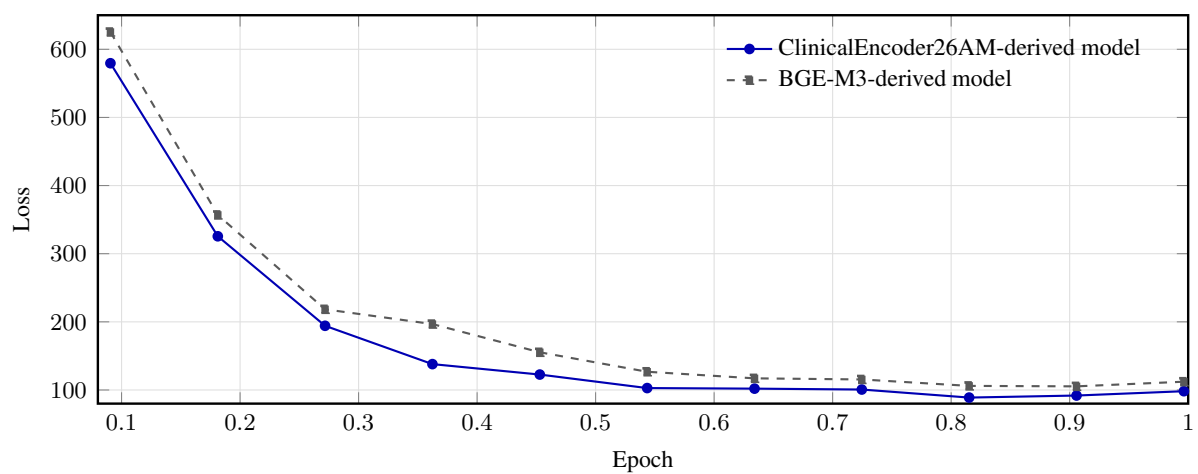
\begin{figure*}[t]
\centering
\begin{tikzpicture}
\begin{axis}[
    width=\linewidth,
    height=0.42\linewidth,
    xlabel={Epoch},
    ylabel={Loss},
    xmin=0.08,
    xmax=1.00,
    ymin=80,
    ymax=650,
    grid=both,
    major grid style={gray!25},
    minor grid style={gray!15},
    legend style={draw=none, fill=none, font=\small},
    legend cell align={left},
    legend pos=north east,
    tick label style={font=\small},
    label style={font=\small},
    line width=0.9pt,
]
\addplot[
    color=blue!70!black,
    thick,
    mark=*,
    mark size=1.6pt,
] coordinates {
    (0.0906,579.5365)
    (0.1812,325.5687)
    (0.2717,194.2215)
    (0.3623,137.9701)
    (0.4529,122.6525)
    (0.5435,102.8171)
    (0.6341,101.9777)
    (0.7246,100.6724)
    (0.8152,88.8760)
    (0.9058,91.8658)
    (0.9964,98.2115)
};
\addlegendentry{ClinicalEncoder26AM-derived model}

\addplot[
    color=black!65,
    thick,
    dashed,
    mark=square*,
    mark size=1.4pt,
] coordinates {
    (0.0906,625.3634)
    (0.1812,356.8888)
    (0.2717,218.4090)
    (0.3623,196.7429)
    (0.4529,155.3395)
    (0.5435,126.7580)
    (0.6341,117.0937)
    (0.7246,115.4627)
    (0.8152,106.0137)
    (0.9058,105.3420)
    (0.9964,112.1600)
};
\addlegendentry{BGE-M3-derived model}
\end{axis}
\end{tikzpicture}
\caption{Training loss over the first epoch for the ClinicalEncoder26AM-derived and BGE-M3-derived MultiClinNER taggers. The clinical post-training initialization converges faster and remains at lower loss through most of finetuning.}
\label{fig:appendix-multiclinner-loss}
\end{figure*}

\newpage
\paragraph{Results.}
Our MultiClinNER submission scored exceptionally well, finishing in the Top 5 across nearly all languages, entity types, and metrics; but the system was clearly recall-oriented.
It ranked first on character-weighted recall in most languages across all entity types, which indicates that the clinically post-trained encoder localized relevant evidence broadly and transferred well across languages, including Czech.
Its weaker side was precision, especially on strict metrics, where long or clinically compositional mentions were sometimes broken into multiple nearly adjacent spans.
Our ablation study demonstrated the usefulness of the pre-training, as the clinically post-trained model converged faster and stayed at lower loss than the base BGE-M3 model during downstream finetuning, as shown in Figure~\ref{fig:appendix-multiclinner-loss}.
This supports the view that the post-training step does not merely improve retrieval-style semantic matching, but also provides a better initialization for downstream span-level and token-level information extraction.

\paragraph{Reusability.}
In the present paper, we reuse these predictions only as auxiliary evidence during CA26AM decoding.
Concretely, they help bridge difficult cases where the alignment scores identify the right region but leave small gaps between salient target tokens.
The orthogonality of the two approaches made the system useful as an auxiliary source of "inside-a-span" signal for CA26AM decoding, but not strong enough to replace alignment as the primary transfer mechanism.

\paragraph{Conclusion.}
In the MultiClinNER shared task, ClinicalEncoder26AM demonstrated that clinically grounded multilingual token representations can transfer effectively to clinical named entity recognition with only a lightweight supervised head.
Its strongest pattern was broad multilingual recall: the model consistently localized clinically relevant evidence across languages, even when precision and exact boundary control remained less strong.
This profile suggests a pragmatic division of labor.
The clinically post-trained encoder provides strong semantic coverage, while explicit span boundaries are learned by a lightweight downstream head.
That is also why these predictions remain useful in the present pipeline: we do not use them as the primary transfer mechanism, but as complementary evidence in cases where alignment identifies the right target region yet leaves locally fragmented support that would benefit from merging.

\end{document}